\documentclass[10pt,twocolumn,letterpaper]{article}

\usepackage{iccv}
\usepackage{times}
\usepackage{epsfig}
\usepackage{graphicx}
\usepackage{amsmath}
\usepackage{amssymb}
\usepackage{multirow}
\usepackage[british,UKenglish,USenglish,english,american]{babel}
\usepackage{tabularx}
\usepackage{authblk}
\usepackage{url}

\usepackage[breaklinks=true,bookmarks=false]{hyperref}
\newcommand{\Tref}[1]{Table~\ref{#1}}
\newcommand{\Eref}[1]{Equation~(\ref{#1})}
\newcommand{\Fref}[1]{Figure~\ref{#1}}
\newcommand{\Sref}[1]{Section~\ref{#1}}

\iccvfinalcopy 

\def\iccvPaperID{****} 

\graphicspath{ {images/} }
\begin{document}

\title{Coherent Online Video Style Transfer}

\author[1]{Dongdong Chen\thanks{This work was done when Dongdong Chen is an intern at MSR Asia.}}
\author[2]{Jing Liao}
\author[2]{Lu Yuan}
\author[1]{Nenghai Yu}
\author[2]{Gang Hua}
\affil[1]{University of Science and Technology of China, \authorcr \tt\small {cd722522@mail.ustc.edu.cn, ynh@ustc.edu.cn}}
\affil[2]{Microsoft Research Asia, \authorcr \tt\small{\{jliao,luyuan,ganghua\}@microsoft.com }}

\maketitle

\begin{abstract}
Training a feed-forward network for fast neural style transfer of images is proven to be successful. However, the naive extension to process video frame by frame is prone to producing flickering results. We propose the first end-to-end network for online video style transfer, which generates temporally coherent stylized video sequences in near real-time. Two key ideas include an efficient network by incorporating short-term coherence, and propagating short-term coherence to long-term, which ensures the consistency over larger period of time. Our network can incorporate different image stylization networks. We show that the proposed method clearly outperforms the per-frame baseline both qualitatively and quantitatively. Moreover, it can achieve visually comparable coherence to optimization-based video style transfer, but is three orders of magnitudes faster in runtime.

\end{abstract}

\section{Introduction}

\begin{figure*}[ht]
	\centering
	\includegraphics[width=0.99\textwidth]{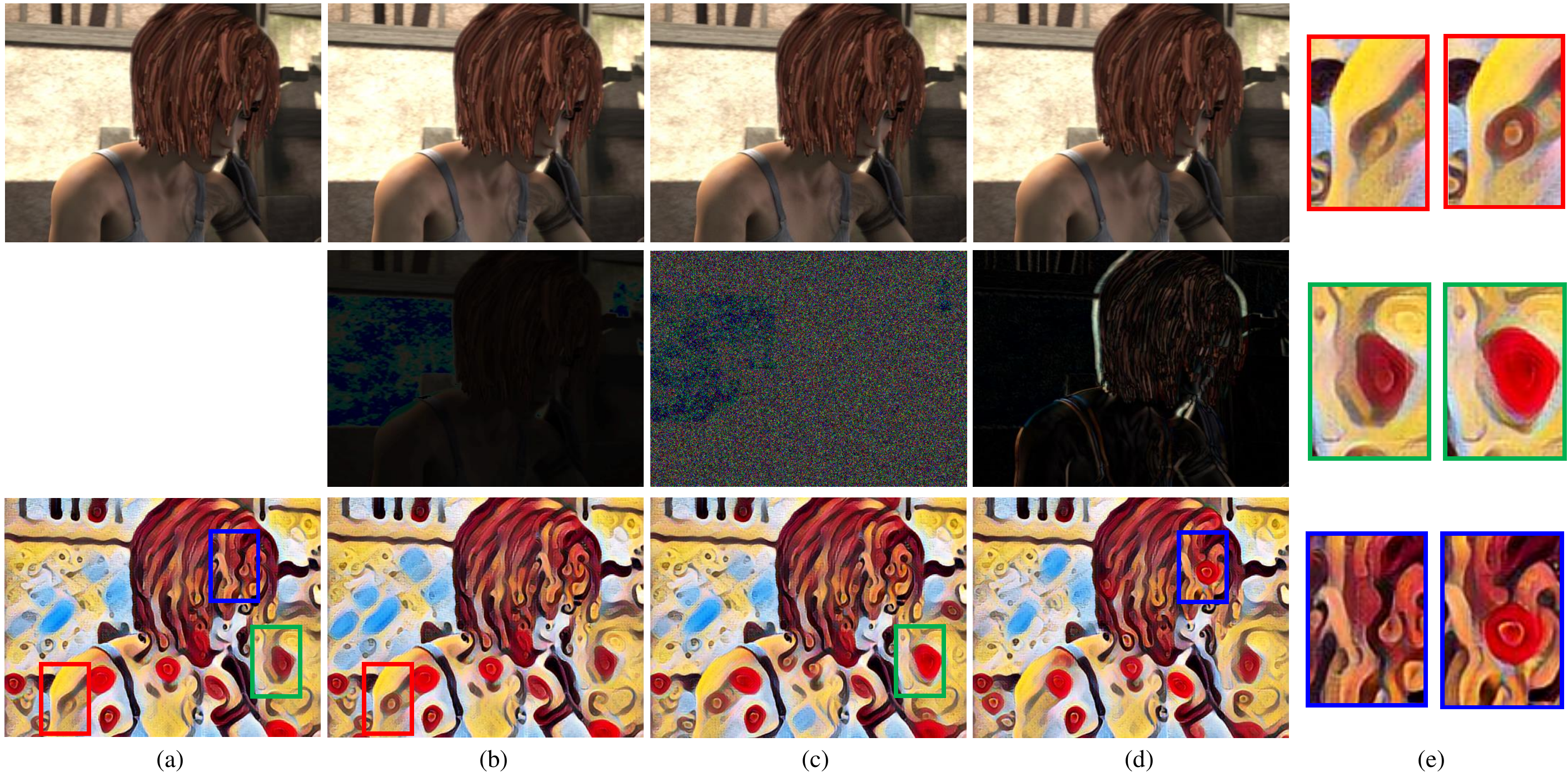}
	\caption{The image stylization network (\eg,~\cite{johnson2016perceptual}), will amplify image variances caused by some unnoticeable changes in inputs. Upper row shows the four inputs: (a) the original one, (b) 5\% lighter than (a), (c) Gaussian noises ($\mu=0, \sigma = 1e-4 $) added to (a); and (d) the next frame of (a) with subtle motions. The middle rows show the absolute difference between (a) and other three inputs. For better visualization, these differences are boosted by $3\times$. The bottom row shows the corresponding stylization results. (e) shows close-up views of some flickering regions.}
	\label{fg:motivation}\vspace{-0.5em}
\end{figure*}

Inspired by the success of work from Gatys et al.~\cite{gatys2015neural} on neural style transfer, there have been a surge of recent works \cite{selim2016painting,chuanli2016,champandard2016semantic,gatys2016controlling} addressing the problem of style transfer using deep neural networks.

In their approaches, style transfer is formulated as an optimization problem, \ie, starting with white noise, searching for a new image presenting similar neural activations as the content image and similar feature correlations as the style image. Notwithstanding their impressive results, these methods are very slow in runtime due to the heavy iterative optimization process. To mitigate this issue, many works have sought to speed up the transfer by training feed-forward networks \cite{johnson2016perceptual,ulyanov2016texture,li2016precomputed,dong2017stylebank,dumoulin2016learned, li2017Diversified}. Such techniques have been successfully applied to a number of popular apps such as Prisma, Pikazo, DeepArt, \etc.

Extending neural style transfer form image to video may produce new and impressive effects, whose appeal is especially strong in short videos sharing, live-view effects, and movie entertainments. The approaches discussed above, when naively extended to process each frame of the video one-by-one, often lead to flickering and false discontinuities. This is because the solution of the style transfer task is not stable. For optimization-based methods (\eg,~\cite{gatys2015neural}), the instability stems from the random initialization and local minima of the style loss function. And for those methods based on feed-forward networks (\eg,~\cite{johnson2016perceptual}), a small perturbation in the content images, \eg, lighting, noises and motions may cause large variations in the stylized results, as shown in~\Fref{fg:motivation}. Consequently, it is essential to explore temporal consistency in videos for stable outputs.

Anderson et al.~\cite{anderson2016deepmovie} and Ruder et al.~\cite{ruder2016artistic} address the problem of flickers in the optimization-based method by introducing optical flow to constrain both the initialization and the loss function. Although very impressive and smoothing stylized video sequences are obtained, their runtime is quite slow (usually several minutes per frame), making it less practical in real-world applications.

In search for a fast and yet stable solution to video style transfer, we present the first feed-forward network leveraging temporal information for video style transfer, which is able to produce consistent and stable stylized video sequences in near real-time. Our network architecture is constituted by a series of the same networks, which considers two-frame temporal coherence. The basic network incorporates two sub-networks, namely the flow sub-network and the mask sub-network, into a certain intermediate layer of a pre-trained stylization network (\eg,~\cite{johnson2016perceptual, dong2017stylebank}).

The flow sub-network, which is motivated by~\cite{zhu2016deep}, estimates dense feature correspondences between consecutive frames. It helps all consistent points along the motion trajectory be aligned in feature domain. The mask sub-network identifies the occlusion or motion discontinuity regions. It helps adaptively blend feature maps from previous frames and the current frame to avoid ghosting artifacts. The entire architecture is trained end-to-end, and minimizes a new loss function, jointly considering stylization and temporal coherence.

There are two kinds of temporal consistency in videos, as mentioned in~\cite{ruder2016artistic}: long-term consistency and short-term consistency. Long-term consistency is more appealing since it produces stable results over larger periods of time, and even can enforce consistency of the synthesized frames before and after the occlusion. This constraint can be easily enforced in optimization-based methods~\cite{ruder2016artistic}. Unfortunately, it is quite difficult to incroporate it in feed-forward networks, due to limited batch size, computation time and cache memory. Therefore, short-term consistency seems to be more affordable by feed-forward network in practice.

Therefore, our solution is a kind of compromise between consistency and efficiency. Our network is designed to mainly consider short-term relationship (only two frames), but the long-term consistency is partially achieved by propagating the short-term ones. Our network may directly leverage the composite features obtained from the previous frame, and combine it with features at the current frame for the propagation. In this way, when the point can be traced along motion trajectories, the feature can be propagated until the tracks end.

This approximation may suffer from shifting errors in propagation, and inconsistency before and after the occlusion. Nevertheless, in practice, we do not observe obvious ghosting or flickering artifacts through our online method, which is necessary in many real applications. In summary, our proposed video style transfer network is unique in the following aspects:\vspace{-0.3em}
\begin{itemize}
  \item Our network is the first network leveraging temporal information that is trained end-to-end for video style transfer, which successfully generates stable results. \vspace{-0.3em}
  \item Our feed-forward network is about thousands of times faster compared to optimization-based style transfer in videos~\cite{anderson2016deepmovie,ruder2016artistic}, reaching 15 fps on modern GPUs.\vspace{-0.3em}
  \item Our method enables online processing, and is cheap in both learning and inference, since we achieve the good approximation of long-term temporal coherence by propagating short-term one.\vspace{-0.3em}
  \item Our network is general, and successfully applied to several existing image stylization networks, including per-style-per-network~\cite{johnson2016perceptual} or mutiple-style-per-network~\cite{dong2017stylebank}.\vspace{-0.3em}
\end{itemize}

\section{Related Work}

\subsection{Style Transfer for Images and Videos}

Traditional image stylization work mainly focus on texture synthesis based on low-level features, which uses non-parametric sampling of pixels or patches in given source texture images~\cite{efros1999texture,hertzmann2001image,efros2001image} or stroke databases~\cite{litwinowicz1997processing,hertzmann1998painterly}. Their extension to video mostly uses optical flow to constrain the temporal coherence of sampling~\cite{bousseau2007video,hays2004image,lu2010interactive}. A comprehensive survey can be found in~\cite{kyprianidis2013state}.

Recently, with the development of deep learning, using neural networks for stylization becomes an active topic. Gatys et al. \cite{gatys2015neural} first propose a method of using pre-trained Deep Convolutional Neural Networks (CNN) for image stylization. It generates more impressive results compared to traditional methods because CNN provides more semantic representations of styles. To further improve the transfer quality, different complementary schemes have been proposed, including face constraints \cite{selim2016painting}, Markov Random Field (MRF) prior~\cite{chuanli2016}, user guidance~\cite{champandard2016semantic} or controls~\cite{gatys2016controlling}. Unfortunately, these methods based on an iterative optimization are computationally expensive in run-time, which imposes a big limitation in real applications. To make the run-time more efficient, some work directly learn a feed-forward generative network for a specific style~\cite{johnson2016perceptual,ulyanov2016texture,li2016precomputed} or multiple styles~\cite{dong2017stylebank,dumoulin2016learned, li2017Diversified} which are hundreds of times faster than optimization-based methods.

Another direction of neural style transfer~\cite{gatys2015neural} is to extend it to videos. Naive solution that independently processes each frame produces flickers and false discontinuities. To preserve temporal consistency, Alexander et al.~\cite{anderson2016deepmovie} use optical flow to initialize the style transfer optimization, and incorporate flow explicitly into the loss function. To further reduce ghosting artifacts at the boundaries and occluded regions, Ruder et al.~\cite{ruder2016artistic} introduce masks to filter out the flow with low confidences in the loss function. This allows to generate consistent and stable stylized video sequences, even in cases with large motion and strong occlusions. Notwithstanding their demonstrated success in video style transfer, it is very slow due to the iterative optimization. Feed-forward networks~\cite{johnson2016perceptual,ulyanov2016texture,li2016precomputed,dong2017stylebank,dumoulin2016learned, li2017Diversified} have proven to be efficient in image style transfer. However, we are not aware of any work that trains a feed-forward network that explicitly takes temporal coherence into consideration in video style transfer.

\subsection{Temporal Coherence in Video Filter}

Video style transfer can be viewed as applying one kind of artistic filter on videos. How to preserve the temporal coherence is essential and has been considered in previous video filtering work. One popular solution is to temporally smooth filter parameters. For instance, Bonneel et al.~\cite{bonneel2013example} and Wang et al.~\cite{wang2006effective} transfer the color grade of one video to another by temporally filtering the color transfer functions.

Another solution is to extend the filter from 2D to 3D. Paris et al.~\cite{paris2011local} extend the Gaussian kernel in bilateral filtering and mean-shift clustering to the temporal domain for many applications of videos. Lang et al.~\cite{lang2012practical} also extend the notion of smoothing to the temporal domain by exploiting optical flow and revisit optimization-based techniques such as motion estimation and colorization. These temporal smoothing and 3D extension methods are specific to their applications, and cannot generalize to other applications, such as stylization.

A more general solution considering temporal coherence is to incorporate a post-processing step which is blind to filters. Dong et al.~\cite{dong2015region} segment each frame into several regions and spatiotemporally adjust the enhancement (produced by unknown image filters) of regions of different frames; Bonneel et al.~\cite{bonneel2015blind} filter videos along motion paths using a temporal edge-preserving filter. Unfortunately, these post-processing methods fracture texture patterns, or introduce ghosting artifacts when applied to the stylization results due to high demand of optical flow.

As for stylization, previous methods (including traditional ones~\cite{bousseau2007video,hays2004image,lu2010interactive,zhang2011online} and neural ones~\cite{anderson2016deepmovie,ruder2016artistic}) rely on optical flow to track motions and keep coherence in color and texture patterns along the motion trajectories. Nevertheless, how to add flow constraints to feed-forward stylization networks has not been investigated before.

\subsection{Flow Estimation}

Optical flow is known as an essential component in many video tasks. It has been studied for decades and numerous approaches has been proposed~\cite{horn1981determining,brox2004high,weickert2006survey,brox2011large,weinzaepfel2013deepflow,revaud2015epicflow}). These methods are all hand-crafted, which are difficult to be integrated in and jointly trained in our end-to-end network.

Recently, deep learning has been explored to solving optical flow. FlowNet~\cite{fischer2015flownet} is the first deep CNNs designed to directly estimate the optical flow and achieve good results. Later, its successors focused on accelerating the flow estimation \cite{ranjan2016optical}, or achieving better quality~\cite{ilg2016flownet}. Zhu et al. \cite{zhu2016deep} recently integrate the FlowNet~\cite{fischer2015flownet} to image recognition networks and train the network end-to-end for fast video recognition. Our work is inspired by their idea of applying FlowNet to existing networks. However, the stylization task, different from the recognition one, requires some new factors to be considered in network designing, such as the loss function, and feature composition, \etc.


\section{Method}

\begin{figure}[t]
	\centering
	\includegraphics[width=1.0\linewidth]{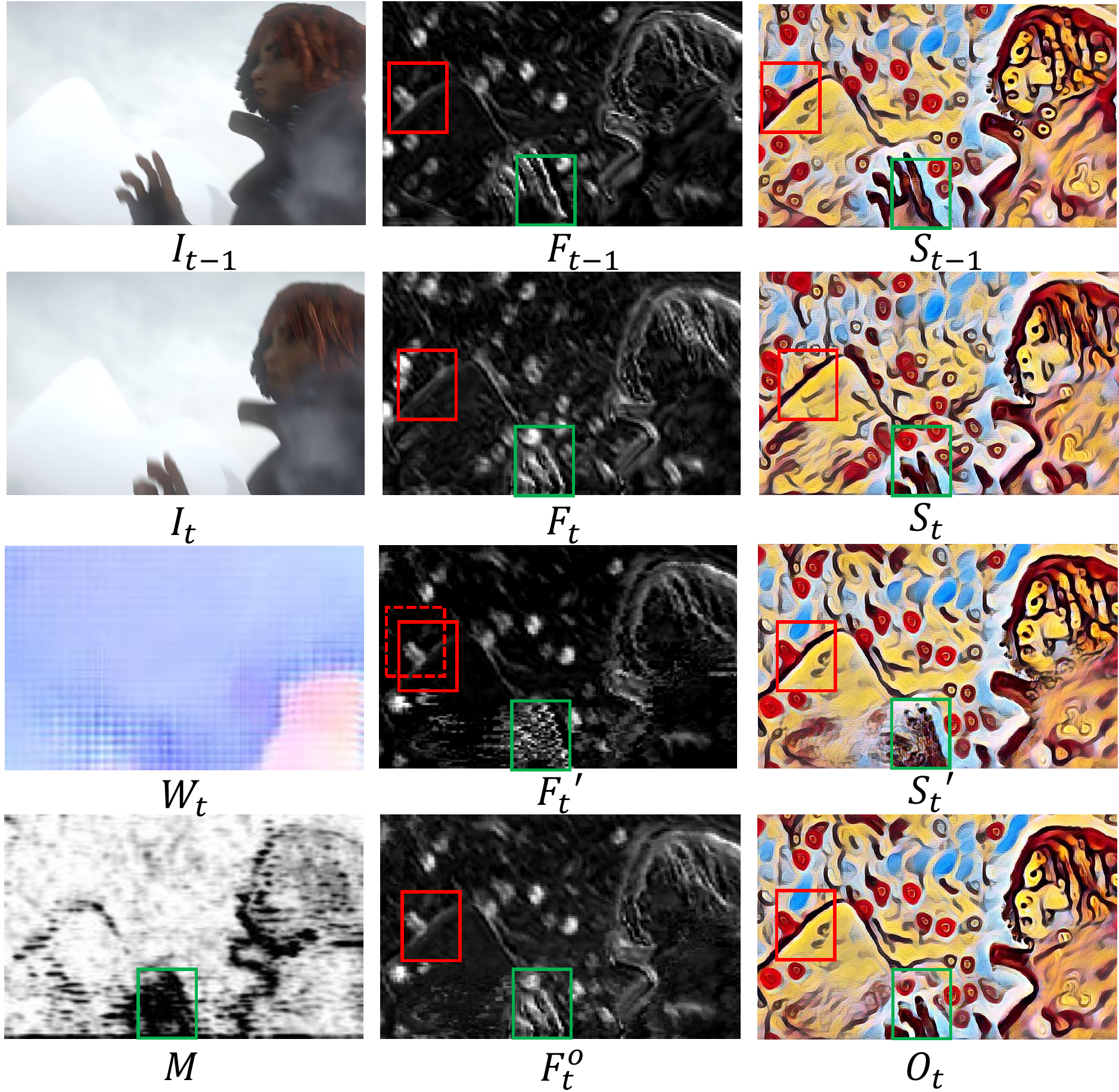}
	\caption{Visualization of two-frame temporal consistency. Two inputs $I_{t-1}, I_{t}$ pass the stylization network~\cite{johnson2016perceptual} to obtain feature maps $F_{t-1}, F_{t}$, and stylized results $S_{t-1}, S_{t}$. We may notice discontinuities between $S_{t-1}$ and $S_{t}$ in \emph{Red} and \emph{Green} rectangles. The third row shows the warped feature map $F_{t}'$ and styled result $S_{t}'$ through flow $W_t$. We can see texture patterns (\emph{Red} rectangle) are successfully traced from $t-1$ to $t$, but ghosting occurs at occluded regions (\emph{Green} rectangle) of $S_{t}'$. The occlusion mask is shown as $M$. In these false regions, $F_{t}'$ (also $S_{t}'$) is replaced with $F_{t}$ (also $S_{t}$) to get the composite features $F^o_{t}$ and result $O_{t}$. }
	\label{fg:intuition}\vspace{-1.0em}
\end{figure}

\subsection{Motivation}

When the style transfer for consecutive frames is applied independently (\eg,~\cite{johnson2016perceptual}), subtle changes in appearance (\eg, lighting, noise, motion) would result in strong flickering, as shown in~\Fref{fg:motivation}. By contrast, in still-image style transfer, such small changes in the content image, especially on flat regions, may be necessary to generate spatially rich and varied stylized patterns, making the result more impressive. Thus, how to keep such spatially rich and interesting texture patterns, while preserving the temporal consistency in videos is worthy of a more careful study.

For simplicity, we start by exploring temporal coherence between two frames. Our intuition is to warp the stylized result from the previous frame to the current one, and adaptively fuse both together. In other words, some traceable points/regions from the previous frame keep unchanged, while some untraceable points/regions use new results occurring at the current frame. Such an intuitive strategy strikes two birds in one stone: 1) it makes sure stylized results along the motion paths to be as stable as possible; 2) it avoids ghosting artifacts for occlusions or motion discontinuities. We show the intuitive idea in~\Fref{fg:intuition}.

The strategy outlined above only preserves the short-term consistency, which can be formulated as the problem of propagation and composition. The issue of propagation relies on good and robust motion estimation. Instead of optical flow, we are more inclined to estimate flow on deep features, similar to~\cite{zhu2016deep}, which may neglect noise and small appearance variations and hence lead to more stable motion estimation. This is crucial to generate stable stylization videos, since we desire appearance in stylized video frames not to be changed due to such variations. The issue of composition is also considered in the feature domain instead of pixel domain, since it can further avoid seam artifacts.

\begin{figure}[t]
	\centering
	\includegraphics[width=1.0\linewidth]{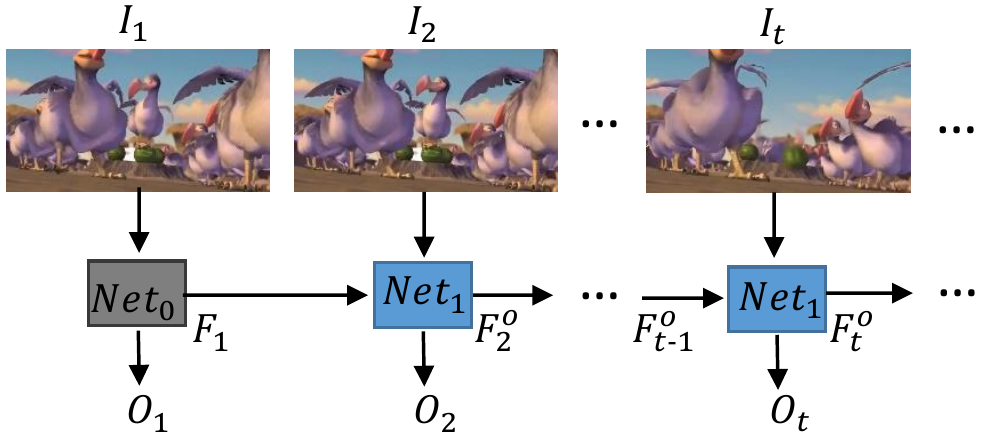}
	\caption{system overview.}
	\label{fg:net_overview}\vspace{-1.0em}
\end{figure}

To further obtain the consistency over long periods of time, we seek a new architecture to propagate short-term consistency to long-term. The pipeline is shown in~\Fref{fg:net_overview}. At $t-1$, we obtain the composite feature maps $F^o_{t-1}$, which are constrained by two-frame consistency. At $t$, we reuse $F^o_{t-1}$ for propagation and composition. By doing so, we expect all traceable points to be propagated as far as possible in the entire video. Once the points are occluded or the tracking get lost, the composite features will keep values independently computed at the current frame. In this way, our network only needs to consider two frames every time, but still approaches long-term consistency.

\begin{figure*}[t]
	\centering
	\includegraphics[width=0.99\textwidth]{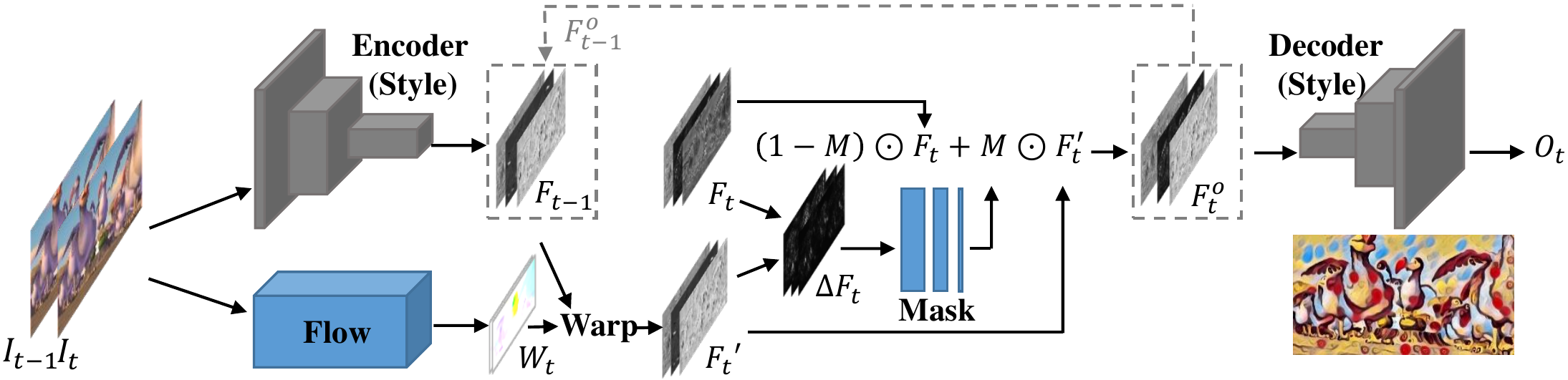}
	\caption{Our network architecture consists of three main components: the pretrained style sub-network, which is split into two parts: an encoder and a decoder; the flow sub-network to predict intermediate feature flow; and the mask sub-network to regress the composition mask.}
	\label{fg:net_structure}
\end{figure*}

\subsection{Network Architecture}

In this section, we explain the details of our proposed end-to-end network for video style transfer. Given the input video sequence $\{I_{t}|t=1...n\}$, the task is to obtain the stylized video sequence $\{O_{t}|t=1...n\}$. The overall system pipeline is shown in~\Fref{fg:net_overview}. At the first frame $I_1$, it uses existing stylization network (\eg,~\cite{johnson2016perceptual}) denoted as $Net_0$ to produce the stylized result. Meanwhile, it also generates the encoded features $F_1$ as the input of our proposed network $Net_1$ at the second frame $I_2$. The process is iterated over the entire video sequence. Starting from the second frame $I_2$, we use $Net_1$ rather than $Net_0$ for style transfer.

The proposed network structure $Net_1$ incorporating two-frame temporal coherence is presented in~\Fref{fg:net_structure}. It consists of three main components: the style sub-network, the flow sub-network, and the mask sub-network. \vspace{-0.8em}

\paragraph{Style Sub-network.} We adopt the pre-trained image style transfer network of Johnson et al.~\cite{johnson2016perceptual} as our default style sub-network, since it is often adopted as the basic network structure for many follow-up work (\eg,~\cite{dumoulin2016learned, dong2017stylebank}). This kind of network looks like auto-encoder architecture, with some strided convolution layers as the encoder and fractionally strided convolution layers as the decoder, respectively. Such architectures allow us to insert the flow sub-network and the mask sub-network between the encoder and the decoder. In Section~\ref{sect:ablation}, we provide the detailed analysis on which layer is better for the integration of our sub-networks. \vspace{-0.8em}

\paragraph{Flow Sub-network.} As a part for temporal coherence, the flow sub-network is designed to estimate the correspondences between two consecutive frames $I_{t-1}$ and $I_{t}$, and then warp the convolutional features. We adopt \emph{FlowNet} (the "Simple" version)~\cite{fischer2015flownet} as our flow sub-network by default. It is pre-trained on the synthetic \emph{Flying Chairs} dataset~\cite{fischer2015flownet} for optical flow task, and should be fine-tuned to produce feature flow suitable for our task.

The process is similar to~\cite{zhu2016deep}, which uses it for video recognition. Two consecutive frames $I_{t-1}, I_{t}$ are firstly encoded into feature maps $F_{t-1}, F_t$ respectively by the encoder. $W_t$ is the feature flow generated by the flow sub-network and bilinearly resized to the same spatial resolution as $F_{t-1}$. As the values of $W_t$ are in general fractional, we warp $F_{t-1}$ to $F_{t}'$ via bilinear interpolation:
\begin{equation}\label{eq:feat_warp}
F_{t}' = \mathcal{W}_{t-1}^t(F_{t-1})
\end{equation}
where $\mathcal{W}_{t-1}^t(\cdot)$ denotes the function that warps features from $t-1$ to $t$ using the estimated flow field $W_t$, namely $F_{t}'(p) = F_{t-1}(p + W_t(p))$, where $p$ denotes spatial location in feature map and flow. \vspace{-0.8em}

\paragraph{Mask Sub-network.} Given the warped feature $F_t'$ and the original feature $F_t$, the mask sub-network is employed to regress the composition mask $M$, which is then adopted to compose both features $F_t'$ and $F_t$. The value of $M$ varies from 0 to 1. For traceable points/regions by the flow (\eg, static background), the value in the mask $M$ tends to be 1. It suggests that the warped feature $F_t'$ should be reused so as to keep coherence. On the contrary, at occlusion or false flow points/regions, the value in the mask $M$ is 0, which suggests $F_t$ should be adopted. The mask sub-network architecture consists of three convolutional layers with stride one. Its input is the absolute difference of two feature maps
\begin{equation}
\Delta F_t = |F_t - F_t'|,
\end{equation}
and the output is a single channel mask $M$, which means all feature channels would share the same mask in the later composition. Here, we obtain the composite features $F_t^o$ by linear combination of $F_t$ and $F_t'$:
\begin{equation}\label{eq:calc_out}
\begin{aligned}
F_t^o &= (1-M) \odot F_t + M \odot F_t'
\end{aligned}
\end{equation}
where $\odot$ represents element-wise multiplication. \vspace{-0.8em}

\paragraph{Summary of \it{$\bf{Net_1}$}.}~\Fref{fg:net_structure} summarizes our network $Net_1$ designed for two frames. Given two input frame $I_{t-1}, I_t$, they are fed into the encoder of fixed style sub-network, generating convolutional feature maps $F_{t-1}, F_t$. This first step is different in inference, where $F_{t-1}$ will not be computed from $I_{t-1}$, and instead borrowed from the obtained composite features $F_{t-1}^o$ at $t-1$. It is illustrated by the dot lines in~\Fref{fg:net_structure}. On the other branch, both frames $I_{t-1}, I_t$ are fed into the flow sub-network to compute feature flow $W_t$, which warps the features $F_{t-1}$ ($F_{t-1}^o$ used in inference instead) to $F_{t}'$. Next, the difference $\Delta F_t$ between $F_{t}$ and $F_{t}'$ is fed into the mask sub-network, generating the mask $M$. New features $F_t^o$ are achieved by linear combination of $F_t$ and $F_t'$ weighted by the mask $M$. Finally, $F_t^o$ is fed into the decoder of the style sub-network, generating the stylized result $O_t$ at frame $t$. For the inference, $F_t^o$ is also the output for the next frame $t+1$. Since both flow and mask sub-networks learn relative flow $W_t$ and mask $M_t$ between any two frames, it is not necessary for our training to incorporate historic information (\eg, $F_{t-1}^o$) as well as the inference. It can make our training be simple.

\subsection{The Loss Function}

To train both the flow and mask sub-networks, we define the loss function by enforcing three terms: the coherence term $\mathcal{L}_{cohe}$, the occlusion term $\mathcal{L}_{occ}$, and the flow term $\mathcal{L}_{flow}$. The coherence term $\mathcal{L}_{cohe}$ penalizes the inconsistencies between stylized results of two consecutive frames.
\begin{equation}\label{eq:loss_cohe}
\mathcal{L}_{cohe}(O_t,S_{t-1}) = M^g \odot ||  O_t-\mathcal{W}_{t-1}^t(S_{t-1})||^2,
\end{equation}
where $S_{t-1}$ is the stylized result produced independently at $t-1$. The warping function $\mathcal{W}_{t-1}^t(\cdot)$ uses the ground-truth flow $W_t^g$. $M^g$ is the ground-truth mask, where $1$ represents consistent points/regions and $0$ represents untraceable ones. It encourages the stylized result $O_t$ to be consistent with $S_{t-1}$ in the traceable points/regions.

On the contrary, in the untraceable regions (e.g. occlusions), the occlusion term $\mathcal{L}_{occ}$ enforces $O_t$ to be close to the independently stylized result $S_{t}$ at frame $I_t$:
\begin{equation}\label{eq:loss_func_occ}
\mathcal{L}_{occ}(O_t,S_{t}) = (1-M^g)\odot||O_t- S_t||^{2}.
\end{equation}

Besides, we add a term to constrain the feature flow:
\begin{equation}\label{eq:loss_func_flow}
\mathcal{L}_{flow} = ||W_t - W_t^g\downarrow||^{2}.
\end{equation}
Here we use the down-scaled version of the ground-truth optical flow $W_t^g\downarrow$, which is re-scaled to the same size of $W_t$, to serve as the guidance for feature flow estimation.

In summary, our loss function to train flow and mask sub-networks is the weighted avearge of three terms.
\begin{equation}\label{eq:loss_func}
\mathcal{L} = \alpha\mathcal{L}_{cohe} + \beta\mathcal{L}_{occ} + \lambda\mathcal{L}_{flow},
\end{equation}
where $\alpha = 1e5$, $\beta = 2e4$ and $\lambda = 20$ by default.

Note that our loss function discards the content and style loss for training the original style network, because the pre-trained style sub-network is fixed during the training period of the flow and mask sub-networks. We believe that $S_t$ (or $S_{t-1}$) itself can provide sufficient style supervision in learning. One extra benefit is that we can directly leverage other trained still-image style models and apply it to videos directly. In this sense, our proposed framework is general.

\section{Experiments}

\subsection{Dataset Set-up} \label{sec:data_acq}

Our task requires a big video dataset with varied types of motions and ground-truth optical flow. However, existing datasets are quite small, \eg, the synthetic \emph{MPI Sintel} dataset~\cite{Butler:ECCV:2012} (only has 1,064 frames totally). Instead, we collect ten short videos (eight animation movies episode of \emph{Ice Age}, and two real videos from YouTube), around $28,000$ frames together as our training dataset.

To obtain approximated ground-truth flow $W^g$ between every two consecutive frames in these videos, we use DeepFlow2~\cite{weinzaepfel2013deepflow} to compute the bidirectional optical flow and use the backward flow as the ground-truth.

As for the ground-truth of the composition mask $M^g$, we adopt the methods used in~\cite{ruder2016artistic,sundaram2010dense} to detect occlusions and motion boundaries. We mask out two types of pixels, being set to $0$ in $M^g$: 1) the occlusion pixels achieved by cross-checking the forward and backward flows; 2) the pixels at motion boundaries with large gradients of flow, which are often less accurate and may result in ghosting artifacts in composition. All other pixels in $M^g$ are set to $1$.

We use the \emph{MPI Sintel}~\cite{Butler:ECCV:2012} as the test dataset, which is widely adopted for optical flow evaluation. It contains 23 short videos and is labeled with ground-truth flow and occlusion mask. The dataset covers various types of real scenarios, such as large motions and motion blurs.

\subsection{Implementation details}

In our experiments, we adopt two types of pre-trained style network (per-style-per-net~\cite{johnson2016perceptual}\footnote{In our experiment, we adopt the released model of \cite{johnson2016perceptual}. It uses two stride-2 convolutions to down-scale the input followed by five residual blocks and then two convolutional layers with stride-1/2 to up-scale, but its channel number of all convolutional layers is half of~\cite{johnson2016perceptual}.}, multiple-style-per-net~\cite{dong2017stylebank}\footnote{We slightly modified the StyleBank model~\cite{dong2017stylebank}, whose encoder and decoder sub-networks adopted the same structures as~\cite{johnson2016perceptual}, but the stylebank layer is inserted after the third residual block.}) as our fixed style sub-network. We train the flow sub-network and mask sub-network on the video dataset described in~\Sref{sec:data_acq}. All videos have the image resolutions of $640\times360$. The network is trained with a batch size of 1 (frame pair) for 100k iterations. And the Adam optimization method~\cite{kingma2014adam} is adopted with the initial learning rate of $1e-4$ and decayed by $0.8$ at every $5k$ iterations.

\subsection{Quantitative and Qualitative Comparison}
\begin{table}
\footnotesize
\begin{center}
\setlength\tabcolsep{4.9pt} 
\begin{tabular}{r|c|c|c|c|c}
\hline
\multirow{2}{*}{Methods} & \multicolumn{4}{c|}{stability error $e_{stab}$} & runtime\\
\cline{2-5}
 & \textit{Muse} & \textit{Candy} & \textit{Scream} & \textit{Udnie} & (fps)\\
\hline
Johnson et al. \cite{johnson2016perceptual}& 0.0199 & 0.0240 & 0.0048 & 0.0108  & 38.17 \\
\cite{johnson2016perceptual}+Ours & \textbf{0.0121} & \textbf{0.0105} & \textbf{0.0034} & \textbf{0.0076}  & \textbf{15.07} \\
\cite{johnson2016perceptual}+Ours$\dagger\dagger$ & 0.0135 & 0.0120 & 0.0036 & 0.0079 & 15.07 \\
\hline
Dong et al. \cite{dong2017stylebank} & 0.0159 & 0.0181 & 0.0035 & 0.0059 & 20.6 \\
\cite{dong2017stylebank} + Ours & \textbf{0.0126} & \textbf{0.0131} & \textbf{0.0030} & \textbf{0.0048} & \textbf{7.35} \\
\hline
\end{tabular}
\end{center}
\caption{Comparison of different methods on stability error and runtime (GPU Titan X). Compared to the per-frame processing baseline \cite{johnson2016perceptual} or \cite{dong2017stylebank}, our method can obtain much lower stability loss while only $2.5\sim2.8\times$ slower. Compared to fixed flow sub-network (indicated by $\dagger\dagger$), our fine-tuned flow sub-network achieves better coherence.}
\label{tb:quan_eval}\vspace{-0.7em}

\end{table}

For video style transfer, runtime and temporal consistency are two key criteria. Runtime uses the frame rate of inference. The temporal consistency is measured by
\begin{equation}\label{eq:stability_error}
e_{stab}(O_t,O_{t-1}) = M^g \odot ||  O_t-\mathcal{W}_{t-1}^t(O_{t-1})||^2,
\end{equation}
where the stability error $e_{stab}(O_t,O_{t-1})$ measures the coherence loss (in \Eref{eq:loss_cohe}) between two results $O_t$ and $O_{t-1}$. Here, we only evaluate stability of results on traceable regions. Lower stability error indicates more stable result. For the entire video, we use the average error instead.

\textbf{Quantitative Results.} To validate the effectiveness of our method, we test and compare using two existing stylization networks~\cite{johnson2016perceptual,dong2017stylebank}. The baseline for comparison is to apply their networks to process each frame independently. As shown in~\Tref{tb:quan_eval}, for all the four styles, our method obtains much lower stability error than the baseline~\cite{johnson2016perceptual,dong2017stylebank}. As for the runtime, our method is around $2.5\sim2.8\times$ slower than the baseline, because our network may need extra computation in both flow and mask sub-networks. Nevertheless, our method is still near real-time (15 fps in Titan X).

As a reference, we also test the optimization method~\cite{ruder2016artistic} with the \textit{Candy} style on our test database. Ours is with slightly larger temporal coherence error compared to theirs ($0.0067$), because our network is trained for all videos while theirs is optimized  for one. As for the speed, ours is thousands of times faster than theirs ($0.0089$ fps).

\begin{figure*}[t]
	\centering
	\includegraphics[width=0.95\textwidth]{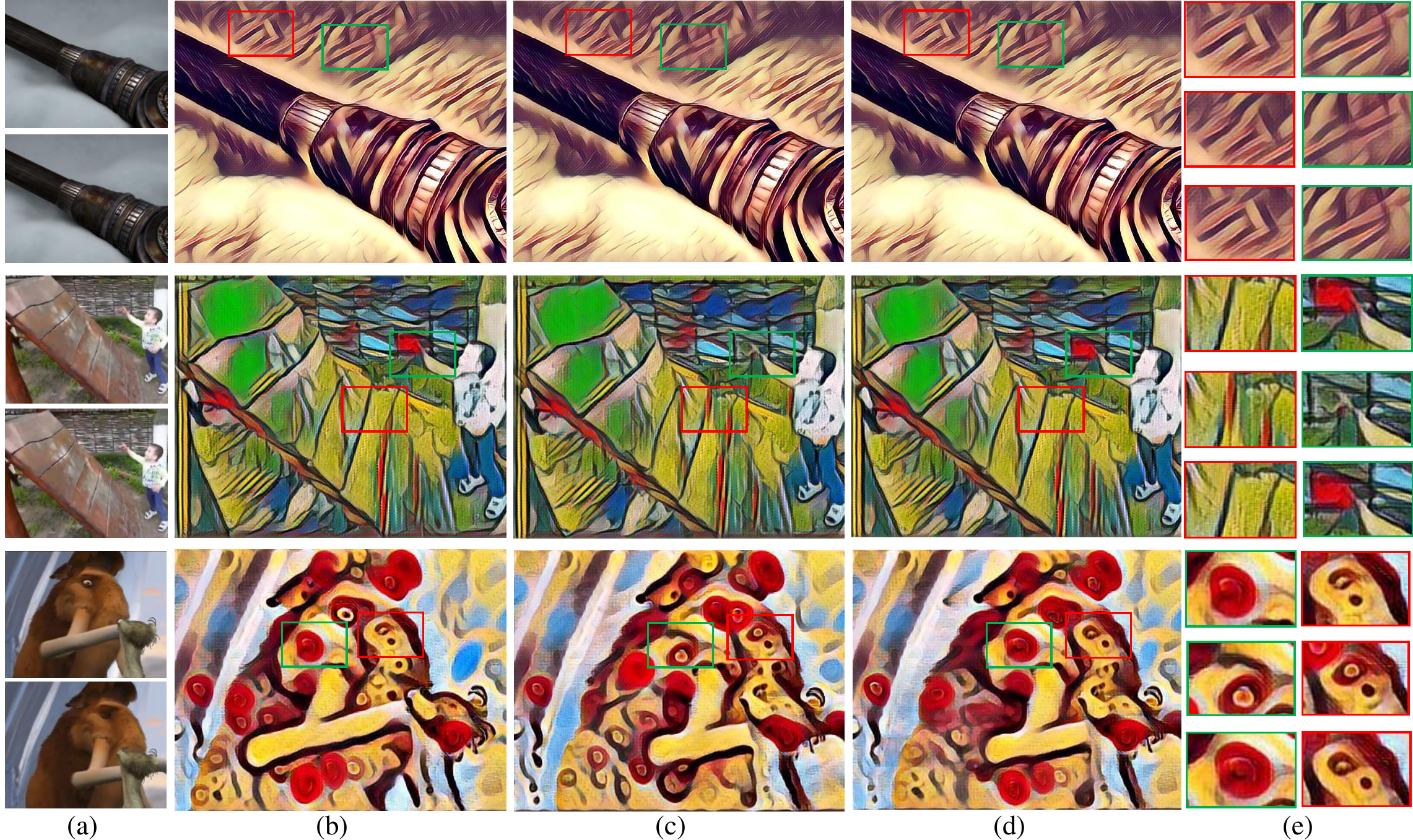}
	\caption{Qualitative comparison results: (a) consecutive frames pair (top: frame $t$, bottom: frame $t+1$); (b) stylization result of frame $t$; (c) stylization result of frame $t+1$ from baseline; (d) stylization result of frame $t+1$ from our method; (e) top to down: dilated marked regions corresponding to (b),(c),(d) respectively. The top row are with \cite{dong2017stylebank} on a nearly static scene, and bottom two rows are with~\cite{johnson2016perceptual} on a scene with camera motion or object motion. Compared to baseline of \cite{johnson2016perceptual,dong2017stylebank}, our results are all more temporal coherent.}
\label{fg:qual_eval}
\end{figure*}

\textbf{Qualitative Results.} In~\Fref{fg:qual_eval}, we show three examples with kinds of representative motions to visually compare our results with per-frame processing models~\cite{dong2017stylebank,johnson2016perceptual}. These results clearly show that our methods successfully reduce temporal inconsistency artifacts which appear in these per-frame models. In the nearly static scene (\emph{First Row}), ours can keep the scene unchanged after stylization while the per-frame models fail. As for the scenes with motions, including both camera motions (\emph{Second Row}) and object motions (\emph{Third Row}), our method keeps the coherence between two frames except for the occluded regions. (\textbf{\emph{The comparisons in our supplementary video \footnote{\url{http://home.ustc.edu.cn/~cd722522/}} are highly recommended for better visualization.}})

\begin{figure}[t]
	\centering
	\includegraphics[width=1.0\linewidth]{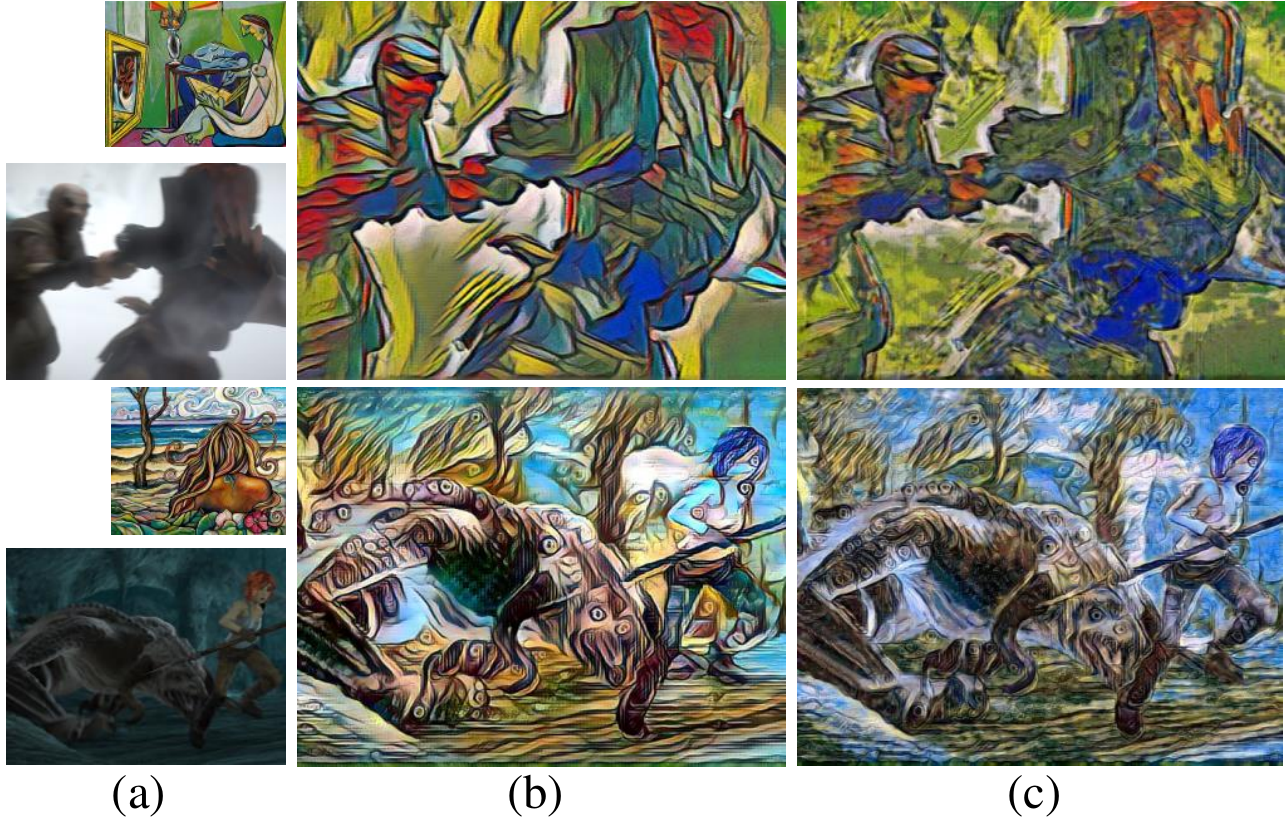}
	\caption{Comparison of our results (b) and results of \cite{bonneel2015blind} (c) on the same inputs (a). Their post-processing scheme results in ghosting and blurring artifacts for video style transfer task.}
	\label{fg:blind_cmp}
\end{figure}

We further compare our method with a post-processing method~\cite{bonneel2015blind}, which is applied to the stylized results produced by per-frame model~\cite{johnson2016perceptual}. As shown in~\Fref{fg:blind_cmp}, the results produced from the post-processing method~\cite{bonneel2015blind} look not so clear as ours, and produces ghosting artifacts. This is because optimizing temporal coherence after stylization may not be able to obtain the global optima for both temporal coherence and stylization.

\subsection{Ablation Study}
\label{sect:ablation}

\textbf{Layer Choice for Feature Composition.} To study which layer of the style sub-network is the best for our feature propagation and composition, we try different layers for integration. For the basic style network~\cite{johnson2016perceptual}, we find 5 intermediate feature layers from input to output (respectively with 1,1/2,1/4,1/2,1 times of original resolution), which allow our flow and mask sub-networks being integrated. The five settings are trained and tested on the same database and with the same style.

In this experiment, we measure the sharpness of their stylization results by \emph{Perceptual Sharpness Index} (PSI)~\cite{feichtenhofer2013perceptual}, in addition to the stability error (in~\Eref{eq:stability_error}). \Tref{fg:blur_cmp} clearly shows that the stability is improved from input to output layers, while the sharpness decreases. It may result from the observation that the stylization networks (\eg,~\cite{johnson2016perceptual}) will amplify the image variances as shown in \Fref{fg:motivation}. When feature flow estimation and composition happen closer to the input layer, small inconsistencies in composite features would also be amplified, causing incoherent results. When they happen closer to the output layer, blending already amplified differences become more difficult and may introduce strong ghosting artifacts. To strike for a balance between stability and image sharpness, we recommend to integrate our sub-networks into the middle layer of stylization networks, \ie, r1/4($E$). In this layer, the image content is compressed as much as possible, which may be beneficial to robust flow estimation and feature composition.

\begin{table}
\small

\begin{tabular}{l|c||l|c}
\hline
 \textit{A} + \textit{B} with ~\cite{johnson2016perceptual} & $e_{stab} $ &  \textit{A} + \textit{B} with ~\cite{dong2017stylebank} & $e_{stab}$\\
\hline
\textit{Scream} + \textit{Scream} & 0.0034 & \textit{Scream} + \textit{Scream} & 0.0031\\
\textit{Scream} + \textit{Candy} & 0.0042 & \textit{Scream} + \textbf{\textit{multiple}} & 0.0032\\
\textit{Candy} + \textit{Scream} & 0.0137 & \textit{Candy} + \textit{Candy} & 0.0121 \\
\textit{Candy} + \textit{Candy} & 0.0105 & \textit{Candy} + \textbf{\textit{multiple}} & 0.0123 \\
\hline
\end{tabular}
\vspace{0.05in}

\caption{Cross comparison of transferring flow and mask sub-networks of \textit{A} to \textit{B}. \textit{A} represents the style of pretrained style sub-network, and \textit{B} is the style which flow and mask sub-networks are trained for. In \cite{dong2017stylebank}, \textit{B} can be multiple styles.}
\label{tb:trans_comp}

\end{table}

\begin{table*}
\small
\centering

\begin{tabularx}{\textwidth}{X|>{\centering}X|>{\centering}X|>{\centering}X|>{\centering}X|>{\centering}X|>{\centering}X}
 &Baseline & r1($E$) & r1/2($E$) & r1/4($E$) & r1/2($D$) & r1($D$) \tabularnewline
    \hline
  $e_{stab}$  & 0.0199 & 0.0187 & 0.0180 & 0.0121 & 0.0058 & 0.0038 \tabularnewline
    \hline
    \textit{ PSI } & 0.4851 & 0.4846 & 0.4839 & 0.4825 & 0.4187 & 0.4086 \tabularnewline
      \hline
\end{tabularx}

\begin{tabularx}{\textwidth}{XXXXXXX}

&
\vspace{0.002em}
   \includegraphics[width=0.12\textwidth]{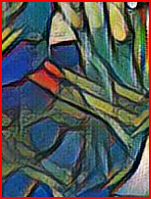}&
   \vspace{0.002em}\includegraphics[width=0.12\textwidth]{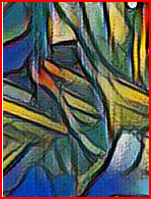}&
   \vspace{0.002em}\includegraphics[width=0.12\textwidth]{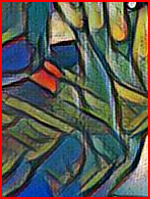}&
    \vspace{0.002em}\includegraphics[width=0.12\textwidth]{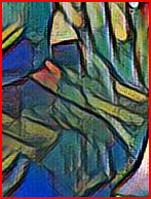}&
   \vspace{0.002em}\includegraphics[width=0.12\textwidth]{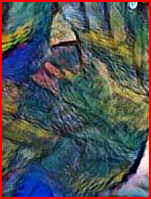}&
   \vspace{0.002em}\includegraphics[width=0.12\textwidth]{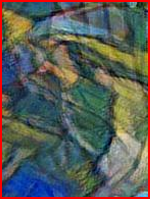}
\end{tabularx}

  \caption{Layer choice for feature composition. r1, r1/2, r1/4 represent different layers whose feature map resolution is $1, 1/2, 1/4 \times$  of the original image, and $E$ and $D$ represent encoder and decoder respectively. The top table shows stability error $e_{stab}$ and PSI for different settings. One visual example is shown on the bottom row.}
  \label{fg:blur_cmp}
\end{table*}

\textbf{Fixed Flow Sub-network.} In our experiment, \textit{FlowNet}~\cite{fischer2015flownet} is adopted as our flow sub-network. Original \textit{Flownet} is trained in image domain for optical flow. It needs to be fine tuned on our task, since the flow would be further improved by jointly learning stylization and temporal coherence. Here, we compare fixed and fine-tuned flow sub-network. As shown in \Tref{tb:quan_eval}, fixed flow sub-network obtains less temporally coherent results than fine-tuned one.

\textbf{Transferability.} To know whether our trained flow and mask sub-networks can be used to a new style (not appearing in training), we conduct two experiments respectively on per-style-per-net~\cite{johnson2016perceptual} and multiple-style-per-net~\cite{dong2017stylebank}. In per-style-per-net~\cite{johnson2016perceptual}, we use two different styles, named as $A$ and $B$ for cross experiments. One combination is style sub-network learned from $A$, and our flow and mask sub-networks learned from $B$. The other combination is reversed. As shown in~\Tref{tb:trans_comp} (\emph{First Column}), it is hard to preserve the original stability when our sub-networks trained on one style are applied to another. By contrast, in multiple-style-per-net~\cite{dong2017stylebank}, our trained sub-networks can be directly used to two new styles without re-training, while preserving the original stability, as shown in~\Tref{tb:trans_comp} (\emph{Second Column}). The observation suggests that our sub-networks learned with multiple-style-per-net~\cite{dong2017stylebank} can be independent of styles, which is beneficial to real applications.

\section{Conclusion and Discussion}

In this paper, we present the first end-to-end training system by incorporating temporal coherence for video style transfer, which can speed up existing optimization-based video style transfer (\cite{anderson2016deepmovie,ruder2016artistic}) by thousands of times, and achieve near real-time speed on modern GPUs. Moreover, our network achieves the long-term temporal coherence through the propagation of the short-term ones, which enables our model for online processing. It can be successfully employed in existing stylization networks~\cite{johnson2016perceptual,dong2017stylebank}, and can even be directly used for new styles without re-training. Our method can produce stable and visually appealing stylized videos in the presence of camera motions, object motions, and occlusions.

There are still some limitations in our method. For instance, limited by the accuracy of ground-truth optical flow (given by DeepFlow2~\cite{weinzaepfel2013deepflow}), our results may suffer from some incoherence where the motion is too large for the flow to track. And after propagation over a long period, small flow errors may accumulate, causing blurriness. These open questions are interesting for further exploration in the future work.

\section*{Acknowledgement}
This work is partially supported by National Natural Science Foundation of China(NSFC, NO.61371192)

{\small
\bibliographystyle{ieee}
\bibliography{egbib}
}

\end{document}